\documentclass{article}
\usepackage{ijcai24}

\usepackage{times}
\usepackage{soul}
\usepackage{url}
\usepackage{tabularx} 
\usepackage{makecell} 
\usepackage{booktabs}

\usepackage[hidelinks]{hyperref}
\usepackage[utf8]{inputenc}
\usepackage[small]{caption}
\usepackage{graphicx}
\usepackage{amsmath}
\usepackage{amsthm}
\usepackage{booktabs}
\usepackage{algorithm}
\usepackage{algorithmic}
\usepackage[switch]{lineno}
\usepackage{inconsolata}
\usepackage{tabularx}
\usepackage{booktabs}
\usepackage{graphicx}
\usepackage{subcaption}

\newcommand\citet[1]{\citeauthor{#1}~[\citeyear{#1}]}
\linenumbers

\title{A Survey of Multimodal Sarcasm Detection}

\author{
    Shafkat Farabi\textsuperscript{1}, Tharindu Ranasinghe\textsuperscript{2}, Diptesh Kanojia\textsuperscript{3}, Yu Kong\textsuperscript{4}, Marcos Zampieri\textsuperscript{1}
    \affiliations
   \textsuperscript{1}George Mason University, USA, \\
   \textsuperscript{2}Lancaster University, UK \\ 
   \textsuperscript{3}University of Surrey, UK, \\
   \textsuperscript{4}Michigan State University, USA
       \emails
    mfarabi@gmu.edu
}

\begin{document}

\maketitle
\nolinenumbers
\begin{abstract}
    Sarcasm is a rhetorical device that is used to convey the opposite of the literal meaning of an utterance. Sarcasm is widely used on social media and other forms of computer-mediated communication motivating the use of computational models to identify it automatically. While the clear majority of approaches to sarcasm detection have been carried out on text only, sarcasm detection often requires additional information present in tonality, facial expression, and contextual images. This has led to the introduction of multimodal models, opening the possibility to detect sarcasm in multiple modalities such as audio, images, text, and video. In this paper, we present the first comprehensive survey on multimodal sarcasm detection - henceforth MSD - to date. We survey papers published between 2018 and 2023 on the topic, and discuss the models and datasets used for this task. We also present future research directions in MSD.  
\end{abstract}

\section{Introduction}
Sarcasm is a sophisticated linguistic phenomenon wherein individuals articulate thoughts using words that convey the opposite of their intended meaning \cite{tiwari-etal-2023-predict}. The Cambridge English Dictionary defines sarcasm as \emph{``The use of remarks that clearly mean the opposite of what they say, made in order to hurt someone's feelings or to criticize something in a humorous way''}. Sarcasm is prevalent in user generated content across social media platforms such as Twitter (now known as X), Facebook, and Reddit, as well as in popular culture, including sitcoms and movies. Many use sarcasm to convey contempt, anger, humor, or derogatory sentiments \cite{maynard-greenwood-2014-cares}.

Sarcasm is often expressed using incongruity between spoken words and the intended sentiment. It frequently relies on usage of hyperbole and reference to contextual world knowledge \cite{chaudhari2017literature}. These strategies make automatically identifying sarcasm a challenging yet interesting task for many applications. For example, the figurative nature of sarcasm makes it an often-quoted challenge for sentiment analysis \cite{joshi2017automatic}. Detection of elements of sarcasm helps to resolve seemingly contradictory sentiments like \emph{``The restaurant was so clean that I could barely avoid stepping into the puddle!''} \cite{badlani-etal-2019-ensemble}. \citet{maynard-greenwood-2014-cares} show that correctly detecting sarcasm can significantly improve sentiment analysis systems. Similarly, sarcasm plays a crucial role in offensive speech and humor identification \cite{frenda2018role}. Finally, applications that model mental health on social media can also benefit from sarcasm identification. \citet{rothermich2021change} show a correlation between sarcasm use and mental conditions such as anxiety and depression.

The importance of sarcasm identification for better understanding communication cannot be overstated. We find growing interest in the problem within the AI, Computer Vision, Speech Processing, and NLP communities that motivates us to present this survey. To the best of our knowledge, this is the first comprehensive survey on MSD filling an important gap in the literature. We survey over 60 papers that present datasets and computational approaches to detect sarcasm and we describe them in detail in this survey. The remainder of this paper is organized as follow: Section \ref{sec:multimodal} discusses text-based sarcasm detection as compared to multimodal approaches. Visuo-Textual detection of sarcasm is discussed in Section 3 while Section 4 discusses Audio-Visual \& Textual detection. Section 5 concludes this survey and presents avenues for future work.

\section{Textual \textit{vs.} Multimodal Sarcasm Detection}
\label{sec:multimodal}

A clear majority of the previous works in automatic sarcasm detection have focused on text classification. Often portrayed as a supervised machine learning problem, several datasets have been introduced for text-based sarcasm detection. The biggest and most popular such dataset, \texttt{SARC} \cite{sarc}, contains 533 million sarcastic and 1.3 million non-sarcastic data collected from Reddit. \texttt{iSarcasm} is another such popular dataset \cite{oprea-magdy-2020-isarcasm} containing 777 sarcastic and 3707 non-sarcastic sentences, all collected from Twitter. Numerous studies have been conducted on these datasets over the years, encompassing both conventional machine learning models in the early stages and more recent deep learning approaches such as transformers \cite{hazarika-etal-2018-cascade,liu-etal-2022-dual}. These datasets and methods on text based sarcasm detection are widely discussed in surveys such as \cite{chaudhari2017literature,joshi2017automatic,verma2021techniques,sys-review,alqahtani2023text}.

\begin{figure}[!ht]
  \begin{subcaptiongroup}
    \centering
    \parbox[b]{.24\textwidth}{%
 
    \includegraphics[width=0.95\linewidth]{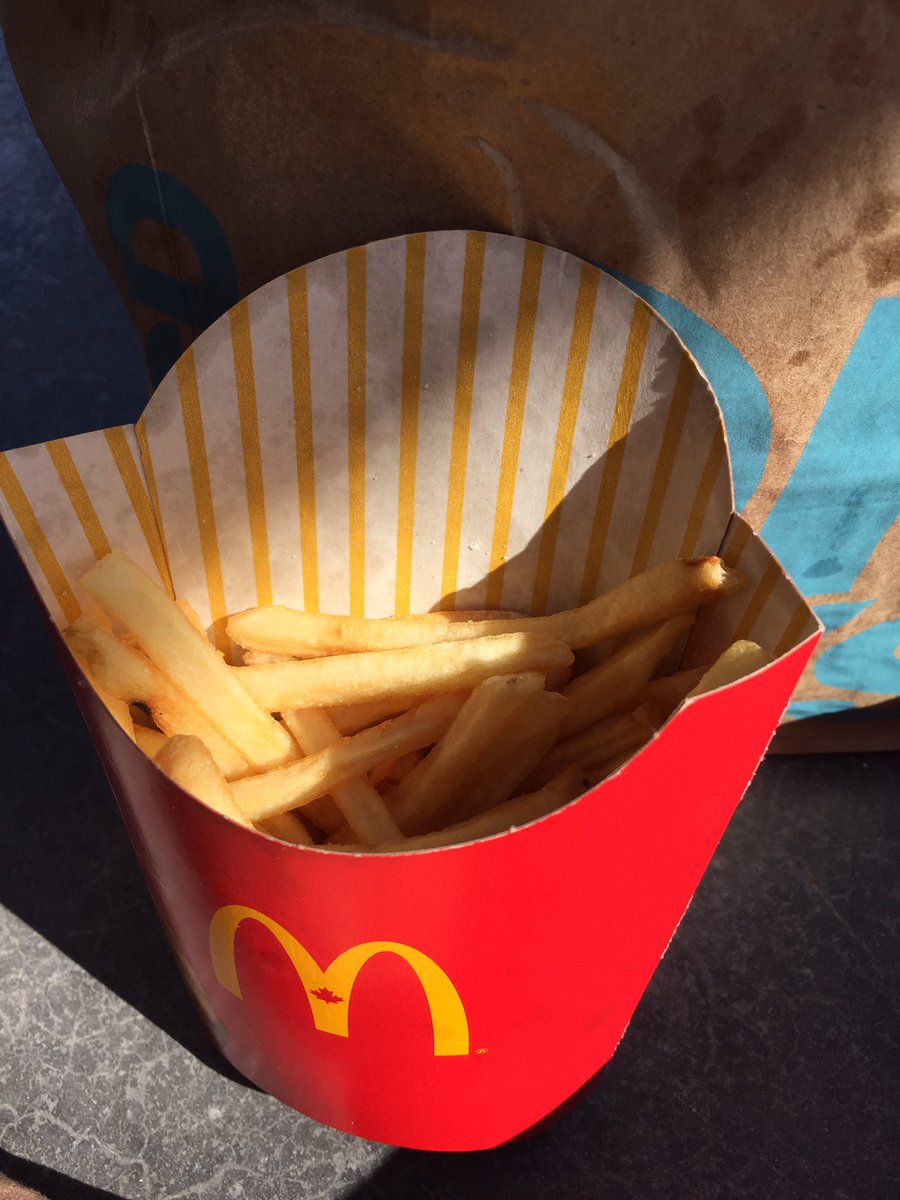}
    \caption{Thanks again for\\ the full fries!}\label{fig:sub1}}%
    \parbox[b]{.24\textwidth}{%
    
    \includegraphics[width=0.95\linewidth]{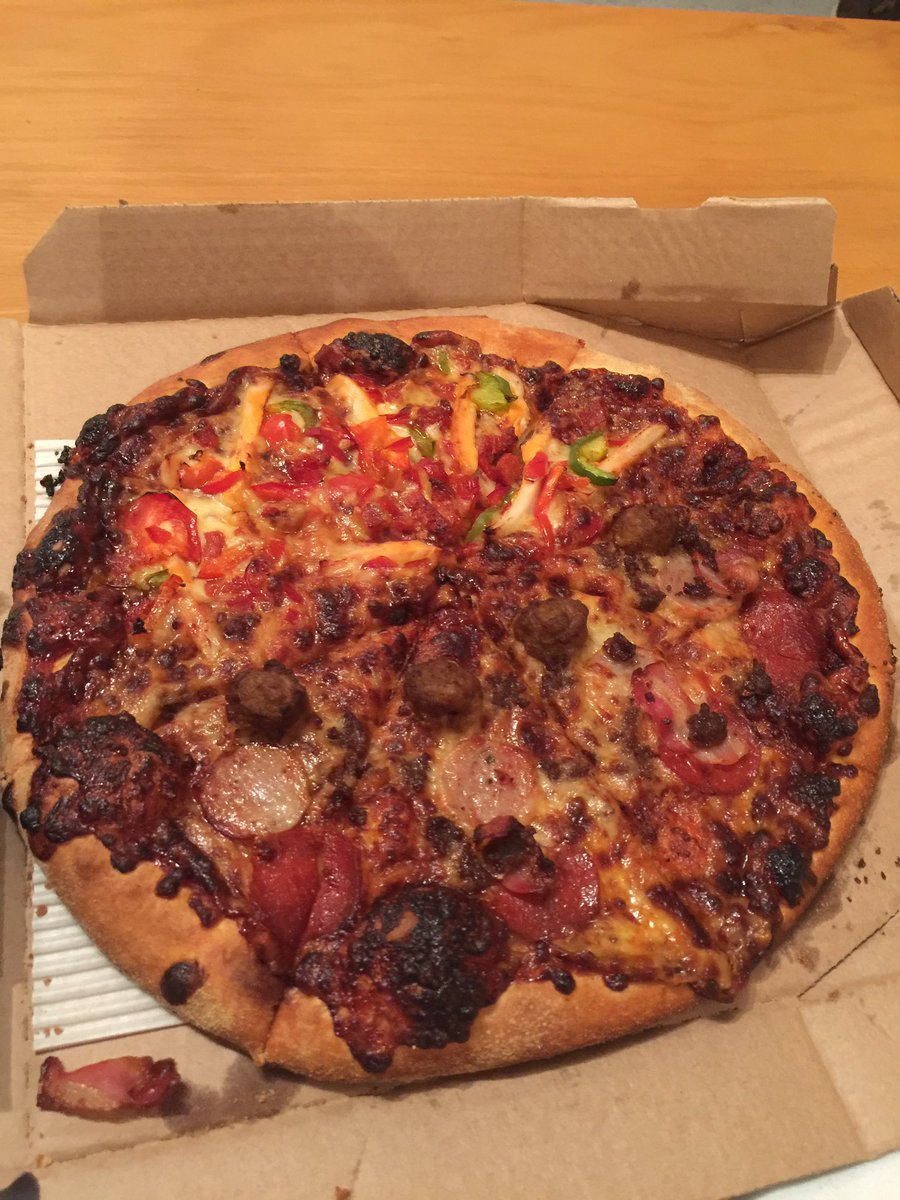}
    \caption{Another perfect pizza\\ from $<$user$>$!}\label{fig:sub2}}%
  \end{subcaptiongroup}
  \caption{Sarcasm using an image accompanying some text.}\label{fig:image1}
\end{figure}

There is yet another form of sarcasm prevalent on social media where users opt for accompanying text with images to express sarcasm. In these cases, the text conveys a meaning that contrasts the content of the image. Figure \ref{fig:image1} presents an example where the text and the image taken individually are not sarcastic but paired together; they express sarcastic intent. 

\begin{figure}[h]
    \centering
    \includegraphics[width=\linewidth]{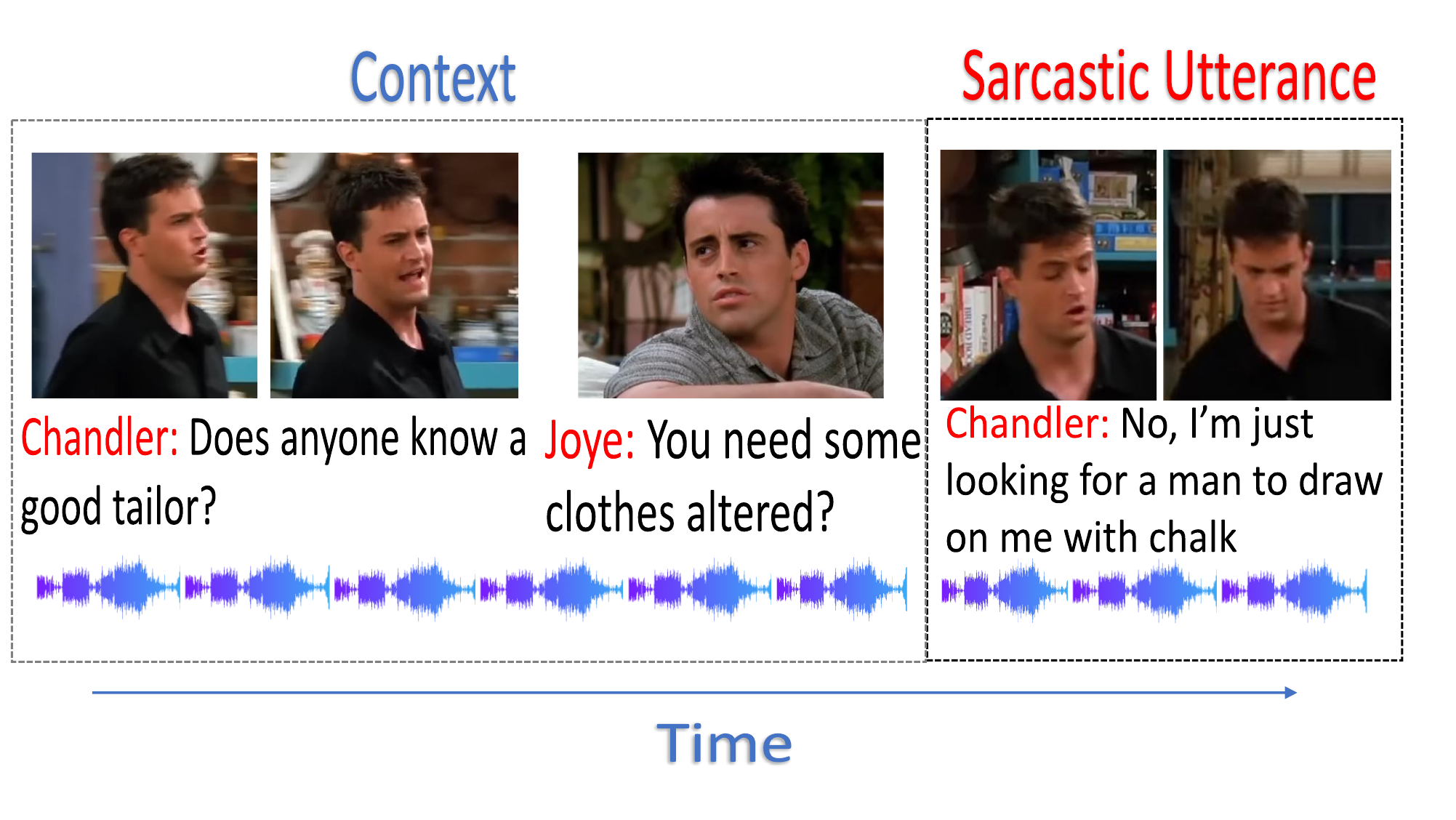}
    \caption{Sarcasm conveyed through text (dialogue), audio (tone), and video (facial expression). The context is important in these cases.}
    \label{fig:image2}
\end{figure}

\noindent Additionally, humans can utilize their facial expressions and voice tone to supplement what they are saying in order to express sarcasm. In this case, video, audio, and text are all necessary to express sarcasm. Figure \ref{fig:image2} shows such a scenario. In this case, the contextual conversation leading up to the sarcastic remark is necessary to appreciate the irony. 

\noindent \citet{cai-etal-2019-multi} pointed out that identifying sarcasm solely based on text is sometimes impossible. In fact, necessary cues to understand sarcasm are often present in the facial cues of the speaker and/or media accompanying the text. Hence, automated models tasked with detecting sarcasm need to be able to take in visual (and sometimes auditory) information to complement the textual data. Following this, there has been an increased effort in the research community to design automatic systems to detect multimodal sarcasm.


Numerous datasets have been curated for MSD with data collected from social media and TV shows such as sitcoms. Curating large human annotated datasets is, however, time consuming and extensive. Accurately annotating sarcastic utterances is a particularly challenging task due to its intrinsic subjectivity. Sarcasm is expressed using underlying incongruity, but this incongruity can be explicitly obvious, or implicitly presented without any negative sentiment phrases; and the degree of incongruity can vary \cite{mishra2016predicting}. Furthermore, the annotator's judgement of sarcasm is also known to be effected by their cultural upbringing \cite{joshi-etal-2016-cultural}.

A multitude of deep learning frameworks have been proposed that can learn from these datasets. Initial works focused on using separate encoders such as ResNets and BERT to encode the data and then proposed novel techniques to fuse these higher-level features. Later works build on top of these approaches by introducing more complex fusion techniques. More recent studies have moved towards an approach of tuning multimodal encoders such as CLIP, VilBERT, and VisualBERT for this specific task. 

\begin{figure}[h]
    \centering
    \includegraphics[width=\linewidth]{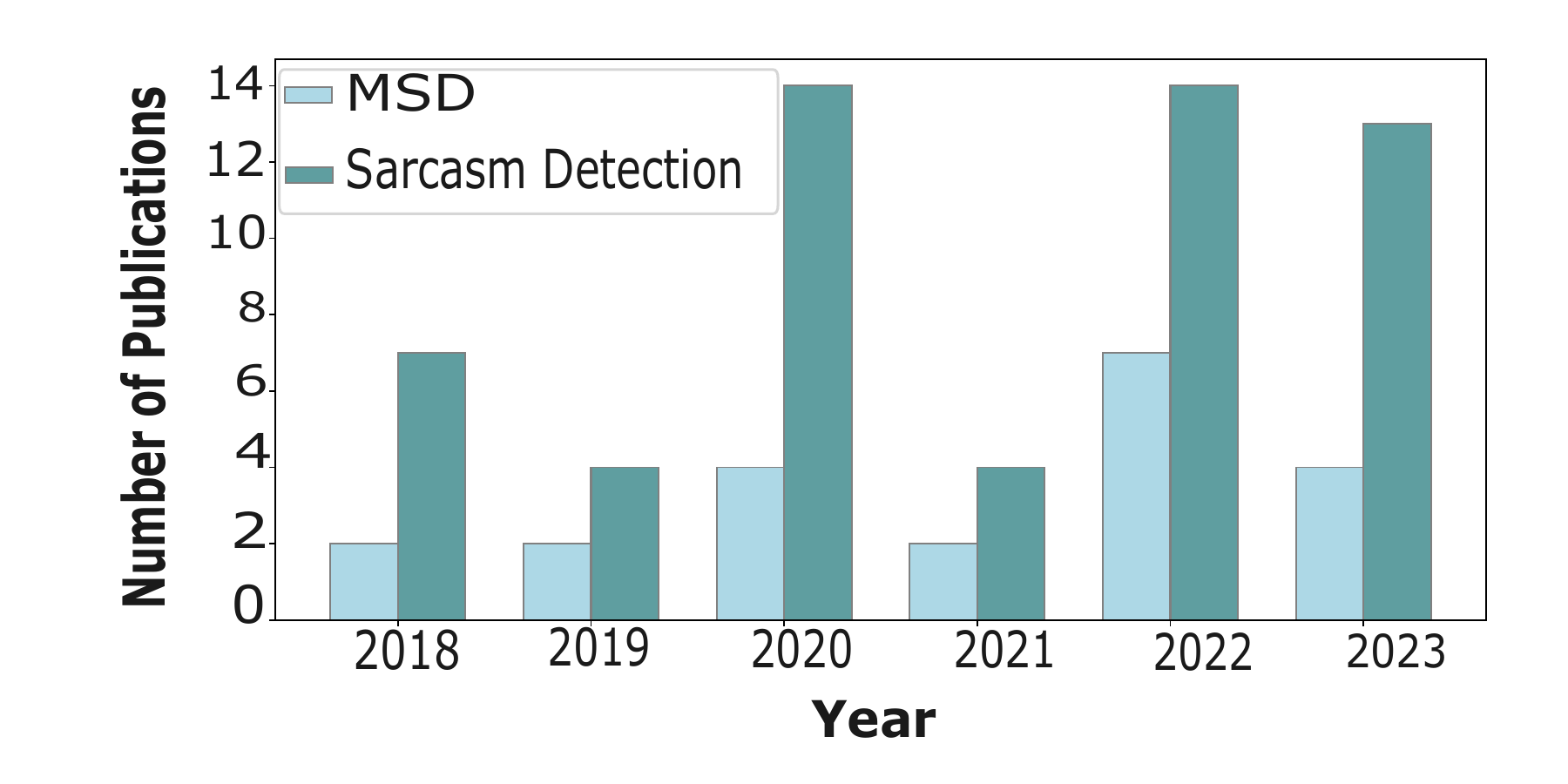}
    \caption{Number of papers on multimodal sarcasm detection and text only sarcasm detection.}
    \label{fig:imagepub}
\end{figure}

\paragraph{Methodology} The previous surveys on sarcasm detection focus solely on text \cite{joshi2017automatic}. In this paper, we fill this important research gap by summarizing the datasets and state-of-the-art methods on MSD in two categories: \textbf{(1) Visuo-Textual and (2) Audio-Visual and Textual Datasets}. In order to find research papers on sarcasm detection, we searched for relevant papers on \emph{Google Scholar} and scientific databases such as ACM Digital Library, IEEE Explore, ACL Anthology, Springer, and CVF Open Access. We use keywords such as \emph{multimodal}, \emph{sarcasm detection}, \emph{sarcasm detection from images}, \emph{social media sarcasm detection}, and others. Most of the papers we report findings from were published in reputed venues such as AAAI, ACL, CVPR, EMNLP, NAACL, and others. We survey papers that curate datasets for MSD or propose a method that can perform substantially well on existing benchmark datasets. In Figure \ref{fig:imagepub}, we present a comparison of the number of studies published in text only sarcasm vs. multi-modal sarcasm detection, between the years 2018-2023. 

\begin{table*}[ht]
\centering
\resizebox{0.95\textwidth}{!}{
\begin{tabular}{@{}lllll@{}}
\toprule
\textbf{Dataset Name} & \textbf{Source} & \makecell[l]{\textbf{Sarcastic} \\ \textbf{Samples}} & \makecell[l]{\textbf{Non-sarcastic} \\ \textbf{Samples}} & \textbf{Additional Remarks} \\
\midrule
\makecell[l]{\texttt{MSDD} \\ \cite{cai-etal-2019-multi}} & Twitter & 10560 & 14075 & \makecell[l]{Tweets containing a picture. \\ Annotated using hashtags in the tweet} \\
\midrule
\makecell[l]{\texttt{MSDD 2.0} \\ \cite{qin-etal-2023-mmsd2}} & Twitter & 11651 & 12980 & \makecell[l]{Enhanced version of MSDD, \\ spurious cues removed \\ manually corrected annotations} \\
\midrule
\makecell[l]{\cite{schifanella2016detecting}} & Instagram, Twitter, Tumbler & 22,025 & 20,025 & Labelled using hashtags, a subset human annotated \\

\midrule
\makecell[l]{\texttt{Silver-Standard Dataset} \\ \cite{ididntmean}} & Instagram & 10,000 & 10,000 & Labelled using hashtags \\
\midrule
\makecell[l]{\texttt{Gold-Standard Dataset} \\ \cite{ididntmean}} & Instagram & 1600 & 10,000 & Human-annotations for sarcasm samples \\
\midrule

\cite{das2018sarcasm} & Facebook & 20,120 & 21,230 & \makecell[l]{Not all samples are multimodal. \\ 98.26\% samples have accompanying images} \\
\midrule
\makecell[l]{\texttt{MORE} \\ \cite{desai2022nice} } & \makecell[l]{\cite{schifanella2016detecting} \\ \& \cite{ididntmean}} & 3510 & - & \makecell[l]{Contains natural language explanation \\ of the sarcasm with non-sarcastic form as well} \\
\bottomrule
\end{tabular}%
}
\caption{Summary of datasets for \textbf{Visuo-Textual} sarcasm detection.}
\label{tab:datasets}
\end{table*}

\noindent We observe that papers on MSD account for significant number of publications on sarcasm detection in this period. Hence, a summary of the literature on MSD is essential to aid future work. We expect this survey to help both researchers already working on MSD as well as researchers new to the task. Furthermore, we believe this is a very timely survey given that the recent introduction of Large Language Models (LLMs) is likely to spark more interest in research on vision and language processing applications.

\section{Visuo-Textual Sarcasm Detection}

The use of images accompanied by text (visuo-textual) are a common way to express sarcasm on social media. Sarcasm expressed through such medium relies heavily on incongruity between the image and text modality. In the following sections, we describe the datasets collected and the approaches to visuo-textual sarcasm detection. 

\subsection{Datasets}

\begin{figure}[!ht]
  
  \begin{subcaptiongroup}
    \centering
    \parbox[b]{.24\textwidth}{%
 
    \includegraphics[width=0.95\linewidth]{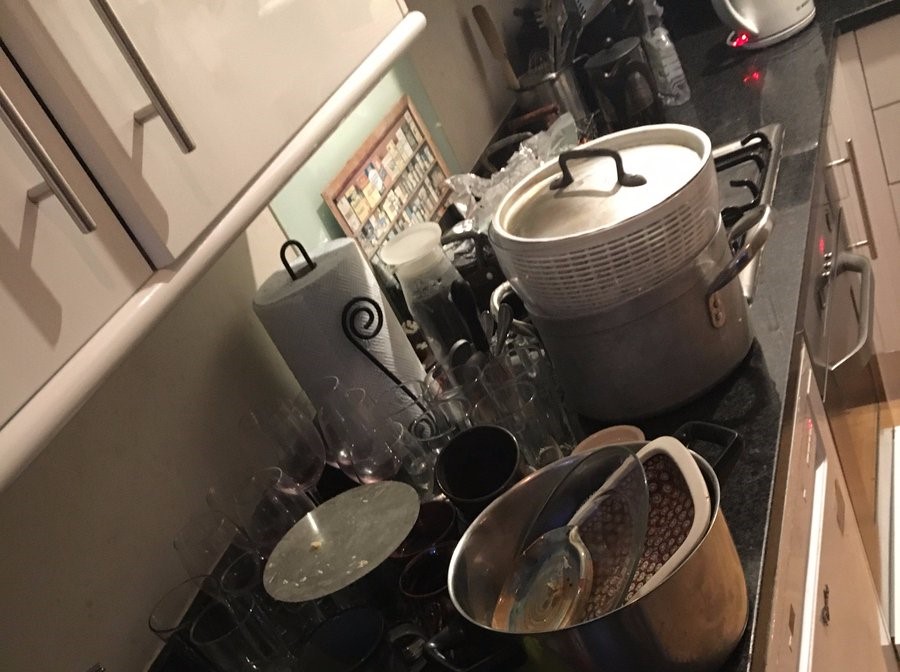}
    \caption{What a joy to wake up the\\ morning after thanksgiving\\ dinner at the hatch flat !!}\label{fig:sub3}}%
    \parbox[b]{.24\textwidth}{%
    
    \includegraphics[width=0.95\linewidth]{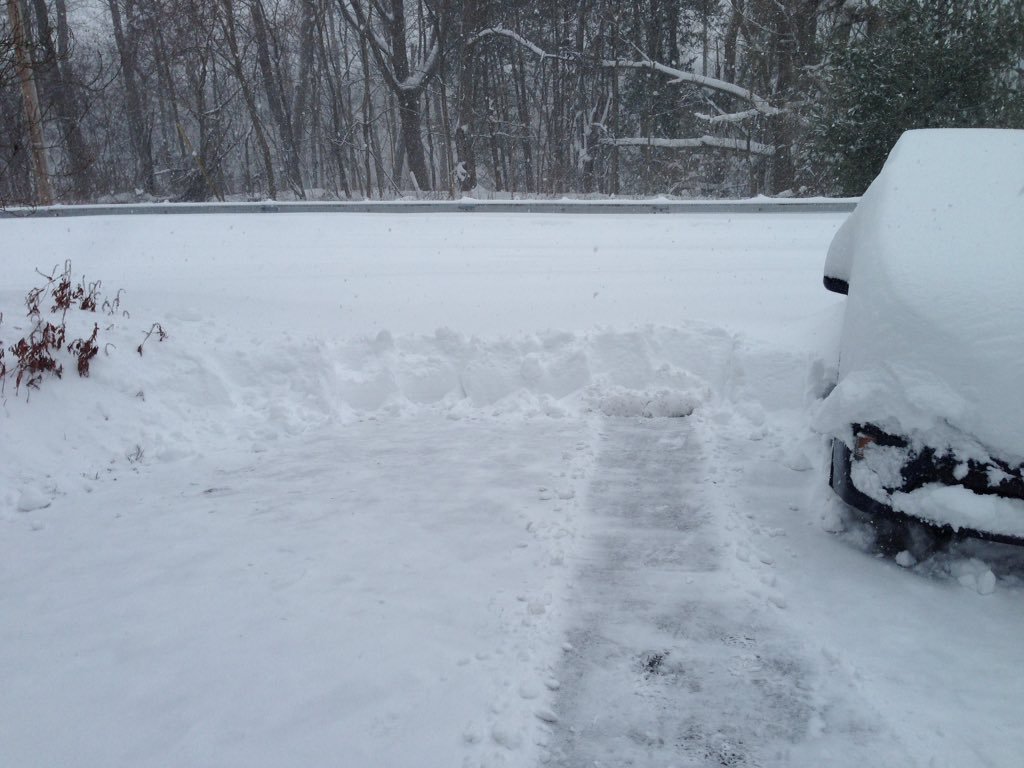}
    \caption{the nice thing is that after i get my driveway shoveled, i can start shoveling the road .}\label{fig:sub4}}%
  \end{subcaptiongroup}
  \caption{Examples from the MMSD \protect \cite{cai-etal-2019-multi} Dataset.}\label{fig:imageMMSD}
\end{figure}

We summarized all datasets for visuo-textual sarcasm detection in Table \ref{tab:datasets}. \citet{schifanella2016detecting} presented one of the first datasets comprised of text and image pairs collected from user posts on Instagram, Twitter, and Tumblr. The authors collect 10,000 sarcastic and 10,000 non-sarcastic posts from Instagram and Tumbler each, and 2,005 sarcastic and 2,005 non-sarcastic posts from Twitter. 
The authors also provide human annotations for 1,000 sarcastic images from Instagram and 1,000 sarcastic images from Tumblr. \citet{cai-etal-2019-multi} presented a similar dataset collected from Twitter. Later works refer to this dataset as the \texttt{MMSD} dataset. We present instances from the \texttt{MMSD} dataset in Figure \ref{fig:imageMMSD}. The dataset contains 19,818 training, 2,410 validation, and 2,409 test examples. Of these, 10,560 samples are sarcastic, and 14,075 non-sarcastic. Tweets containing hashtags similar to \#sarcasm are labelled as sarcastic examples and non-sarcastic otherwise. 

%
%

Later, \citet{qin-etal-2023-mmsd2} enhanced the \texttt{MMSD} dataset by removing spurious cues. They point out that many positive samples in \texttt{MMSD} dataset contain hashtags and emojis that might serve as an easy giveaway to the sarcastic nature of the data. They further point out that some negative samples in \texttt{MMSD} are mis-annotated. They manually re-annotate the negative samples and remove the spurious cues from the positive samples. This version of the dataset is named \texttt{MMSD2.0} and contains 9,572 sarcastic and 10,240 non-sarcastic train samples. The number of sarcastic samples in the validation and test sets changed to 1,042 and 1,037, respectively. The non-sarcastic samples in the validation and test sets changed to 1,368 and 1,372, respectively.

\begin{figure}
  
  \begin{subcaptiongroup}
    \centering
    \parbox[b]{.23\textwidth}{%
 
    \includegraphics[width=0.8\linewidth]{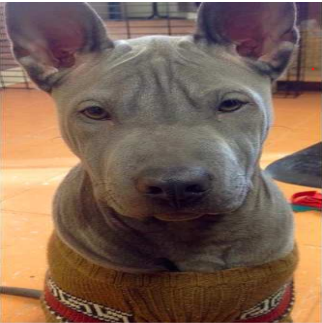}
    \caption{Someone is excited \\for sweater season.}\label{fig:sub3}}%
    \parbox[b]{.23\textwidth}{%
    
    \includegraphics[width=0.8\linewidth]{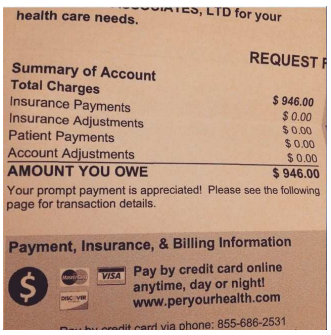}
    \caption{Life is just going great \\ this year so far for me.}\label{fig:sub4}}%
  \end{subcaptiongroup}
  \caption{Example samples from the `Silver-Standard Dataset'.}\label{fig:silver}
\end{figure}

\citet{ididntmean} release a similar dataset composed of image and text pairs collected from Instagram. They release two versions of the dataset of different sizes. The bigger version, named `Silver-Standard Dataset', consists of 20,000 Instagram posts, evenly distributed among positive and negative samples. Similar to \texttt{MMSD}, they determine positive/negative samples based on presence or lack thereof of hashtags in the post (\#sarcasm, \#sarcastic, etc.). They also released a smaller version of the dataset, named `Gold-Standard Dataset', containing only 1,600 human annotated samples. We present examples from this dataset in Figure \ref{fig:silver}.

\begin{table*}[ht]
\centering
\resizebox{0.95\textwidth}{!}{
\begin{tabular}{@{}lcccccccccc@{}}
\toprule
\textbf{Method} & \multicolumn{4}{c}{\textbf{MMSD}} & \multicolumn{4}{c}{\textbf{MMSD 2.0}} & \textbf{Silver} & \textbf{Gold} \\
\cmidrule(r){2-5} \cmidrule(lr){6-9} \cmidrule(l){10-10} \cmidrule(l){11-11}
& acc. & pre. & rec. & f1. & acc. & pre. & rec. & f1. & acc. & acc. \\
\midrule
\cite{cai-etal-2019-multi} & 83.44 & 76.57 & 84.15 & 80.18 & 70.57 & 64.84 & 69.05 & 66.88 & - & - \\
VisualBERT \cite{li2019visualbert}  & 83.51 & 76.66 & 82.94 & 79.68 & - & - & - & - & - & - \\
LXMERT \cite{tan2019lxmert} & 83.93 & 77.83 & 82.59 & 80.14 & - & - & - & - & - & - \\
\cite{xu-etal-2020-reasoning} & 84.02 & 77.97 & 83.42 & 80.60 & - & - & - & - & - & - \\
ViLBERT \cite{10.5555/3454287.3454289}  & 84.68 & 77.52 & 86.37 & 81.71 & - & - & - & - & - & - \\
\cite{pan-etal-2020-modeling} & 86.05 & 80.87 & 85.08 & 82.92 & 80.03 & 76.28 & 77.82 & 77.04 & - & - \\
\cite{10.1145/3474085.3475190} & 86.10 & 85.39 & 85.80 & 85.60 & - & - & - & - & - & - \\
\cite{liang-etal-2022-multi} & 87.55 & 87.02 & 86.97 & 87.00 & 79.83 & 75.82 & 78.01 & 76.90 & - & - \\
\cite{liu-etal-2022-towards-multi-modal} & 87.36 & 81.84 & 86.48 & 84.09 & 76.50 & 73.48 & 71.07 & 72.25\\
\cite{qin-etal-2023-mmsd2} & 88.33 & 82.66 & 88.65 & 85.55 & 85.64 & 80.33 & 88.24 & 84.10 & - & - \\
\cite{wang-etal-2020-building} & 88.51 & 82.95 & 89.39 & 86.05 & - & - & - & - & - & -\\
\cite{pramanick2022multimodal} & 90.82 & - & - & 88.20 & - & - & - & - & - & - \\  
\cite{tian-etal-2023-dynamic} & 93.49 & - & - & 93.21 & - & - & - & - & - & - \\ 
\cite{MMSDPrompt} & 93.85 & 93.57 & 94.89 & 94.06 & - & - & - & - & - & - \\
\cite{ididntmean}  & - & - & - & - & - & - & - & - & 84.22 & 71.5  \\
GPT4 \cite{lin2023goatbench} & 75.88  & - & - & 75.08 & - & - & - & - & - & - \\
InstructBLIP \cite{Yang2023MMBigBenchEM} & 73.10  & - & - & - & - & - & - & - & - & - \\
\bottomrule
\end{tabular}%
}
\caption{A summary of approaches and their performance on \textbf{Visuo-Textual} datasets.}
\label{tab:summary_works}
\end{table*}

\citet{das2018sarcasm} introduced a partially multimodal sarcasm detection dataset collected from Facebook. This dataset contains 20,120 sarcastic and 21,230 non-sarcastic samples, 98.26\% of which include both an image and text. \citet{desai2022nice} investigated a related but novel research field of sarcasm explanation generation from multimodal input. In simple terms, given a sarcastic image-text pair, the goal of this task is to generate a natural language explanation for the pair to be considered sarcastic. To this end, they proposed the \texttt{MORE} dataset containing 3,510 sarcastic utterances, including texts and images. Each utterance is accompanied by a natural language explanation of sarcasm and a non-sarcastic version. 

\subsection{Methods}

The datasets described in the last section have been explored using deep learning approaches. Here we broadly classify these approaches into three classes: (1) traditional deep learning models that use separate encoders for image and text, (2) multimodal transformers, and (3) LLM-based approaches with prompt engineering. We describe the approaches next and summarize their performance on two benchmark datasets in Table \ref{tab:summary_works}.

\paragraph{Traditional deep learning models}  \citet{schifanella2016detecting} were the first to propose that visual contextual features are necessary to decode sarcastic intent from text. Their proposed approach concatenates extracted visual and textual features. Although, the authors demonstrated that visual modality helps in detecting sarcasm in social media, they did not investigate the nature of this relationship, nor did they attempt to engineer specific methods based on the nature of this relation. \citet{das2018sarcasm} utilized CNN from \cite{yfs} to extract image features, and many low level features from post description and user reactions. Their approach can adapt to detecting sarcasm in both text only and image-text multi-modal scenarios in a low-data environment. A key limitation is the dependency on audience reaction. Furthermore, the CNN used for extracting features was noted to incorrectly associate location of the image with sarcastic intent \cite{yfs}. \citet{cai-etal-2019-multi} extracted image attributes and introduced it as a third modality in an attempt to boost performance. The authors identified that multimodal feature fusion by means of simple concatenation is insufficient, and improved this aspect by introducing hierarchical fusion. They performed early fusion by initializing text modality Bi-LSTM with features from the visual modality, and performed representation fusion and modality fusion following the works of \citet{gu-etal-2018-hybrid}. Although the authors attempted to design a sophisticated inter-modal feature fusion technique, they did not try to analyze and take advantage of how information from these modalities interplay with each other.

\citet{pan-etal-2020-modeling} noted that inter and intra modal incongruity play an important role in sarcasm identification. They took advantage of this by proposing a BERT and ResNet based model that can concentrate on both inter-modal and intra-modal incongruity. They accomplished this by introducing attention both in intra-modality and inter-modality fashion. The incongruity between modalities can be disordered and unstructured, which the authors of \citet{pan-etal-2020-modeling} aimed to teach their model solely using data. In order to make this learning process more explicit, \citet{xu-etal-2020-reasoning} proposed the D\&R (Decomposition and Reconstruction) network. They projected the image and text representations in a common subspace, and unique sub-spaces orthogonal to the common space. They later fused features from the unique sub-spaces in an attempt to focus the model more on contrasting elements in modalities. They also extracted adjective-noun pairs (ANP) from the images, and applied ANP aware cross modal attention to make the model more aware of semantic associations between cross modal contexts. To make the learning of inter modal incongruity more structured, \citet{10.1145/3474085.3475190} proposed constructing modality specific and cross-modal dependency graphs from the features extracted through BERT 
and ViT \cite{dosovitskiy2020image}. 
They processed the information stored within these graphs by using interactive graph convolutional networks (GCN). In a later work, \citet{liang-etal-2022-multi} improved further by engineering a method that only focuses on relevant patches of the image that relate to sarcastic cues in the text, achieved by refining the construction of a cross modal graph, by focusing on the objects in the image. More specifically, they followed \citet{8578734} to extract image-attribute pairs from the image. Next, the cross modal graph was generated by using the similarity between image attributes and text words. One drawback of their approach is the dependency on external knowledge to determine word similarity which is a crucial part of their construction of cross modal graph.

The methods discussed thus far focus on using different encoders for different modalities and focus on effectively fusing multi-modal representations in a manner that caters to the incongruity within and between the modalities. Teaching deep networks to find such associations between high level features of different modalities is difficult in a low data environment, even with clever techniques.

\paragraph{Multimodal Transformers} that can encode text and image to a common feature space are becoming popular. \citet{wang-etal-2020-building} benchmarked a few of these methods (VisualBERT, 
LXMERT,
ViLBERT)
for MSD. While a common feature space for multi-modal encodings may help, comparatively smaller pre-training available for these multimodal encoders result to substandard performance compared to uni-modal encoders like BERT and ResNet. However, to show a common feature space for encoding does help, they introduced a trainable bridge between a text-only and image-only encoder, to align them. Along with a 2D intra-attention module for feature fusion, achieving good performance. \citet{qin-etal-2023-mmsd2} presented \texttt{multi-view CLIP}, which further solidifies the idea that a common vision-language feature space facilitates better performance on MSD. They use CLIP \cite{clip}, a popular multi-modal feature extractor, along with clever engineering with transformers for feature fusion.

\begin{table*}[ht]
\centering
\resizebox{0.95\textwidth}{!}{
\begin{tabular}{@{}lllll@{}}
\toprule
\textbf{Dataset Name} & \textbf{Source} & \makecell[l]{\textbf{Sarcastic} \\ \textbf{Samples}} & \makecell[l]{\textbf{Non-sarcastic} \\ \textbf{Samples}} & \textbf{Additional Remarks} \\
\midrule
\makecell[l]{\texttt{MUStARD} \\ \cite{castro2019towards}} & \makecell[l]{TV Shows \\ (YouTube) } & 345 & 345 & \makecell[l]{Includes clips from sitcoms, with contextual data, \\ speaker information. Annotated manually. \\ Non-sarcastic samples collected from MELD \\ \cite{poria-etal-2019-meld}} \\
\midrule
\makecell[l]{\texttt{SE-MUStARD} \\ \cite{chauhan-etal-2020-sentiment}} & \texttt{MUStARD}& 345 & 345 & \makecell[l]{Adds sentiment and emotion labels to \texttt{MUStARD} \\ Annotated manually.} \\
\midrule

\makecell[l]{\texttt{MUStARD++} \\ \cite{ray-etal-2022-multimodal}} & \makecell[l]{TV Shows \\ (YouTube) } & 601 & 601 & \makecell[l]{Enhanced \texttt{MUStARD} with additional videos and labels \\ Provides corrections for some labels in \texttt{MUStARD}.} \\
\midrule
\makecell[l]{\texttt{MUStARD++ Balanced} \\ \cite{bhosale2023sarcasm}} & \makecell[l]{\texttt{MUStARD++} \\ \& House MD }   & 691 & 674 & Extended to balance the sarcasm types \\

\midrule
\makecell[l]{\texttt{SEEmoji MUStARD} \\ \cite{chauhan2022emoji}} & \texttt{MUStARD++} & 691 & 601 & Augmented with emojis, sentiment, and emotions \\

\midrule
\makecell[l]{ Spanish Multimodal Sarcasm \\ \cite{alnajjar-hamalainen-2021-que}} & Archer, South Park & 90 & 869 & Voice and text are in Spanish, manually annotated \\
\bottomrule
\end{tabular}%
}

\caption{Summary of datasets for \textbf{Audio-Visual \& Textual} sarcasm detection.}
\label{tab:datasetsvideo}
\end{table*}
\paragraph{Prompting and LLMs} \citet{MMSDPrompt} explored prompt-tuning for multimodal sarcasm detection. They modeled sarcasm detection as a masked language prediction task and integrated it with a ViT for image encoding and an inter-modality attention transformer to predict the sarcasm level of the text. They used a dot product based similarity assessment similar to the approach of  \citet{clip} for providing supervision for training the model.

\citet{lin2023goatbench} studied the ability of multimodal LLMs to identify social abuse in social media memes in a nuanced fashion. They released a multi-task benchmark (GOAT Benchmark) containing different task categories, one of which is multimodal visuo-textual sarcasm and consists of data sampled from MMSD dataset. They benchmarked several popular LLMs like GPT4 and
LLaVa-1.5 
on these tasks with template prompting. \citet{Yang2023MMBigBenchEM} released another multimodal benchmark for LLMs titled MM-BigBench containing sarcasm detection as a task and samples data from MMSD. They benchmarked various LLMs like GPT4, LLaMa, OpenFlamingo, Blip, and InstructBLIP etc. Despite being initial works touching multimodal sarcasm detection with LLMs, sarcasm was not the primary focus here. Furthermore, they did not experiment with prompt tuning the LLMs to improve detection performance, focusing on LLM capabilities in a zero shot scenario.

\section{Audio-Visual \& Textual Sarcasm Detection}
In this modality, sarcasm is detected via audio and video recordings of dialogue accompanied by text captions. This type of sarcasm is very prevalent in sitcoms, TV shows, and stand-up comedy. While sarcasm can also exist in short video social media platforms such as TikTok and YouTube, all the datasets and methods developed to tackle this modality have focused on sitcoms and TV shows. 

\subsection{Datasets}

A summary of datasets developed for audio-visual and textual sarcasm is presented in Table \ref{tab:datasetsvideo}. The most widely-used dataset is \texttt{MUStARD} \cite{castro2019towards} which contains sarcastic clips from popular sitcom TV shows, namely \emph{Friends}, \emph{The Golden Girls}, \emph{Sarcasmaholics Anonymous}, and \emph{The Big Bang Theory}. For non-sarcastic utterances, the authors reuse data from a multimodal emotion recognition dataset called \texttt{MELD} \cite{poria-etal-2019-meld}. The dataset is balanced, manually annotated and contains 690 samples. Each utterance contains a video clip, audio, and captions in text with necessary contextual conversation leading to the utterance. The context includes audio, video, captions and speaker identifiers, as often these clips contain a conversation between multiple parties. 

Numerous research studies have extended the \texttt{MUStARD} dataset in several directions. \citet{chauhan-etal-2020-sentiment} introduced sentiment and emotion labels in the \texttt{MUStARD} dataset, thereby building \texttt{SE-MUStARD}, showing that these labels improve sarcasm detection. \citet{ray-etal-2022-multimodal} further enhanced the \texttt{SE-MUStARD} dataset to almost double its size and introduce emotion, valence, arousal, and sarcasm-type labels. They also corrected nearly 399 annotation errors in \citet{chauhan-etal-2020-sentiment}'s emotion labels. They published this corrected and enhanced dataset as \texttt{MUStARD++}. \texttt{MUStARD++} is enhanced with 264 new videos from `The Big Bang Theory', and `The Silicon valley'. The total number of sarcastic utterances in this dataset is 601. In order to keep it balanced, additional non-sarcastic examples were also included.

\citet{bhosale2023sarcasm} noticed that the newly introduced `sarcasm types' category  in \texttt{MUStARD++} is imbalanced. In order to address this issue, they augmented the dataset with 90 sarcastic and 74 non-sarcastic samples taken from the TV series `House MD'. They manually annotated the `sarcasm types' label of the newly introduced data and named this extended dataset \texttt{MUStARD++ Balanced}. As a by-product of this effort, they increased the diversity by adding new data. In more recent work, \citet{chauhan2022emoji} published another version  of \texttt{MUStARD} called \texttt{SEEmoji MUStARD}. The authors noted that emotions and sentiments are sometimes implicit and difficult to decipher from text. But emojis can play an important role by alluding to the implicit emotions embedded in the text by the speaker. Hence, they appended appropriate emojis from a pool of 25 most popular ones used in social media, along with the sentiment (positive/negative/neutral) and emotion labels of the emojis to each sample of this dataset. 

Finally, \citet{alnajjar-hamalainen-2021-que} presented a dataset in Spanish comprised of clips taken from \emph{Archer} and \emph{South park} TV shows. The dataset contains 960 utterances, of which only 90 are sarcastic, and 869 are non-sarcastic. The dataset does not contain any train-test splits. The authors include two different dialects of Spanish, and manually annotate samples. 

\subsection{Methods}

\begin{table*}[ht]
\centering

\begin{tabular}{@{}llcccc@{}}
\toprule
\textbf{Method}                          & \textbf{Dataset}          & \textbf{Accuracy} & \textbf{Precision} & \textbf{Recall} & \textbf{F1 Score} \\ \midrule
\cite{castro2019towards}                & \texttt{MUStARD}                   & -                 & 72.6               & 71.6            & 71.6              \\
\cite{chauhan-etal-2020-sentiment}      & \texttt{Se-MUStARD}                & -                 & 73.4               & 72.8            & 72.6              \\
IWAN\cite{9387561}                          & \texttt{MUStARD}                   & -                 & 75.2               & 75.2            & 75.1              \\
\cite{ray-etal-2022-multimodal}         & \texttt{MUStARD}                   & -                 & 74.2               & 74.2            & 74.2              \\

\cite{10235179}                         & \texttt{MUStARD}                   & 79.32             & 78.1               & 77.42           & 77.6              \\
MuLOT \cite{pramanick2022multimodal}     & \texttt{MUStARD}             & 78.57             & -                  & -               & -                 \\
\cite{ray-etal-2022-multimodal}         & \texttt{MUStARD ++}                & -                 & 70.3               & 70.3            & 70.3              \\
\cite{tiwari-etal-2023-predict}         & \texttt{MUStARD ++}                & -                 & 73.2               & 73.2            & 73.3             \\
\cite{bhosale2023sarcasm}         & \texttt{MUStARD ++}                & -                 & 73.5               & 72.8            & 73.1              \\
\cite{bhosale2023sarcasm}         & \texttt{MUStARD ++ Balanced}                & -                 & 73.8               & 73.5            & 73.6              \\
\cite{chauhan2022emoji}                 & \texttt{SEEmoji-MUStARD}           & -                 & 77.9               & 76.9            & 76.7              \\ 
\cite{alnajjar-hamalainen-2021-que}         & \texttt{Spanish Multimodal Sarcasm}                & 93.1                 & -             & -           & -              \\

\bottomrule
\end{tabular}
\caption{A summary of approaches and their performance on \textbf{Audio-Visual and Textual} datasets.}
\label{tab:summary_works1}
\end{table*}

Designing methods to detect sarcasm from audio-visual and textual datasets is more difficult than visuo-textual datasets. These datasets are comprised of video and audio, along with transcript of the audio. Often it contains multiple persons having a conversation. The cues for sarcasm such as  intra-modal incongruity can be manifested in a nuanced manner through facial expression, voice tone, and hand gestures \cite{castro2019towards}. 

A summary of all methods and their performance applied to audio-visual and textual sarcasm detection is presented in Table \ref{tab:summary_works1}. The methods experimenting with these datasets are deep learning based, and  these can be broadly classified into three classes: (1) Traditional Deep Learning Approaches fusing the multimodal features by concatenating them, (2) Approaches using Multimodal Attention for feature fusion, and (3) Approaches using Multi-Task learning where sentiment classification is an auxiliary task.


\paragraph{Traditional deep learning approaches} \citet{castro2019towards} was the first study that demonstrates audio and video can help boost performance on MSD, as can the relevant context. Being an initial work, their proposed framework is rather straight forward, using BERT 
Librosa, and
ResNet-152 
for feature extraction, followed by feature concatenation and prediction using an SVM. \citet{alnajjar-hamalainen-2021-que} took on a similar approach of training an SVM to predict sarcasm from concatenated modality specific features. However, they are the only work dealing with Audio-visual and textual detection of sarcasm in a non-English language (Spanish).  Their study re-affirms the importance of multiple modalities for detecting sarcasm, as suggested by \citet{castro2019towards}. However, a benchmarking of the current state-of-the-art \texttt{MUsTARD} dataset frameworks on their \texttt{Spanish Multimodal Sarcasm} dataset is absent from this work.

Furthermore, both suffer from limitations due to lack of investigations in complex multimodal fusion techniques, without the advantage of multiparty conversation relationships, and the utilization of SVM over neural networks.
 
\paragraph{Multimodal Attention} 
\citet{9387561} proposed a multimodal fusion technique that can identify and use information pertaining to inter-modal incongruities. They proposed IWAN model with a focus on such incongruities in the form of positive spoken words paired with negative tone/facial expression, achieved through an attention-based word level scoring mechanism using features from BERT, ResNet and OpenSmile \cite{10.1145/1873951.1874246}. Notably, this technique improved sarcasm detection but they modeled word-tone level incongruity, and left exploration of contextual incongruities for future. \citet{10235179} filled this gap by proposing the use of multi-headed bimodal attention, targeting  incorporation of multimodal incongruities in a global scenario. More complex methods of modeling cross-modal incongruity were hindered by the size of the MUStARD dataset. To circumvent this limitation, \citet{pramanick2022multimodal} proposed the MuLOT framework, where cross-modal incongruity is learned using optimal-transport, while self-attention is introduced to tackle lack of intra-modal incongruity. 
 
Sarcasm can also be identified by readers' gaze pattern. \citet{tiwari-etal-2023-predict} studied this phenomena by incorporating gaze features for multi-modal sarcasm detection. They collected gaze information for a subset of \texttt{MUStARD++} dataset and designed a framework to predict gaze information from textual utterances, demonstrating gaze features with text, video, and audio, improve task performance on \texttt{MUStARD++} dataset.

Not unlike research in visuo-textual sarcasm detection, the trend is now shifting towards using multimodal transformers. The reason for this preference is the fact that multimodal transformers are more capable of identifying both intra and inter modal dependencies from data. \citet{bhosale2023sarcasm} employed a ViFi-CLIP \cite{hanoonavificlip}, a video-text encoder, to encode the video frames as well as the text in a common representation space. They also used a  Wav2vec 2.0 \cite{baevski2020wav2vec}, a self supervised transformer based speech encoder, fine tuned on speech emotion recognition to encode the audio. 

\paragraph{Multi-Tasking with Auxiliary Sentiment Classification:} \citet{chauhan-etal-2020-sentiment} explored the role of speaker sentiment in sarcasm identification. They augmented the \texttt{MUStARD} dataset with emotion and sentiment labels, used attention for aggregating the features and trained their model in a multi-task learning approach where sentiment classification is the auxiliary task. The complex role of sentiment and emotion in the context of sarcasm detection was also explored by \citet{ray-etal-2022-multimodal}. They introduced the \texttt{MUStARD++} dataset, and utilized a collaborative gating strategy for multimodal feature fusion with an extensive ablation study on the effect of speaker information and the modalities. In a later work, \citet{chauhan2022emoji} explored this further by attaching emojis that often have sentiments contrasting that of the sentence. They proposed an emoji-aware-multi-modal-multitask deep learning framework using emotion and sentiment classification as an auxiliary task and evaluate on \texttt{SEEmoji MUStARD}. These studies demonstrate that auxiliary information pertaining to the speaker emotion and sentiment help in detecting irony and sarcasm. 

\section{Conclusion And Future Directions}

This paper presented the first comprehensive survey of MSD. We presented popular datasets as well as computational approaches used for this task. As the interest on MSD continues to grow, we see the following directions for future research. 

\paragraph{Multilingual datasets} As evidenced in this survey, the bulk of work on MSD is on English data, leaving a critical gap within applications developed for other languages. A notable exception is the work by \citet{alnajjar-hamalainen-2021-que} on Spanish. We hope this survey encourages the creation of larger and more comprehensive multilingual data to aid research on MSD on languages other than English. 

\paragraph{Perspectivism} Identifying sarcasm is a highly subjective task for humans. Different people see sarcasm differently, and this is reflected in dataset annotation. Ground truth labels in annotated MSD datasets are based on annotations by a single person or on the aggregation of multiple annotations. We believe that developing MSD models that consider perspectives from multiple annotators, as in \cite{weerasooriya-etal-2023-vicarious} is a more realistic way of representing the problem and it should be explored in the future. 

\paragraph{Inter-task dependencies} Sarcasm is related to other forms of non-literal language such as humor, and also offensive language and hate speech. The recent HahaCkathon shared task at SemEval \cite{meaney2021semeval}, for example, introduced a dataset annotated with respect to humor and offense. This opens the possibility of exploring inter-task dependencies, and to use multi-task learning where MSD can also be modeled jointly with other related tasks. 

\paragraph{LLMs} The recent introduction of a new generation of LLMs is a promising direction for research in MSD. We believe that models that are able to model image and text (e.g., GPT-4) should be further explored for MSD as they have proven to achieve state-of-the-art performance on multiple vision and language tasks. 

\newpage

\bibliographystyle{named}
\bibliography{ijcai24}

\end{document}